%% file: marionet.tex
\documentclass{article}

\usepackage[nonatbib,final]{neurips_2021}
\usepackage[sort&compress,numbers]{natbib}

\usepackage[utf8]{inputenc} 
\usepackage[T1]{fontenc}    
\usepackage{hyperref}       
\usepackage{url}            
\usepackage{booktabs}       
\usepackage{amsfonts}       
\usepackage{nicefrac}       
\usepackage{microtype}      
\usepackage{xcolor}         
\usepackage{xspace}
\usepackage{overpic}
\usepackage{subfigure}
\usepackage[normalem]{ulem}
\usepackage{wrapfig}
\usepackage{graphicx}
\usepackage{amsmath}

\usepackage{mystyle}

\title{MarioNette: Self-Supervised Sprite Learning}

\author{Dmitriy Smirnov\\
  MIT\\
  \And
  Micha\"el Gharbi\\
  Adobe Research\\
  \And
  Matthew Fisher\\
  Adobe Research\\
  \And
  Vitor Guizilini\\
  Toyota Research Institute
  \And
  Alexei A. Efros\\
  UC Berkeley
  \And
  Justin Solomon\\
  MIT
}

\input{include/_notations}

\begin{document}

\maketitle

\begin{abstract}
\input{include/abstract}
\end{abstract}

\input{include/intro}
\input{include/related}
\input{include/method}
\input{include/results}

\input{include/conclusion}
\input{include/acknowledgements}

\bibliographystyle{abbrvnat}
\bibliography{marionet}

\end{document}

%% file: include/_notations.tex
\newcommand{\notation}[1]{\ensuremath{#1}\xspace}

\newcommand{\Width}{\notation{w}}
\newcommand{\Height}{\notation{h}}

\newcommand{\NumFrames}{\notation{n}}
\newcommand{\NumPatches}{\notation{m}}
\newcommand{\PatchSize}{\notation{k}}
\newcommand{\NumLayers}{\notation{\ell}}
\newcommand{\LatentDim}{\notation{d}}

\newcommand{\Dictionary}{\notation{\mathcal D}}
\newcommand{\Frame}[1]{\notation{I_{#1}}}
\newcommand{\Output}[1]{\notation{O_{#1}}}
\newcommand{\Patch}[1]{\notation{P_{#1}}}

\newcommand{\DictLatent}[1]{\notation{z_{#1}}}
\newcommand{\AnchorFeature}[2]{\notation{a_{#1}^{#2}}}
\newcommand{\AnchorProb}[2]{\notation{p_{#1}^{#2}}}
\newcommand{\Transform}[2]{\notation{{\mathcal{T}_{#1}^{#2}}}}
\newcommand{\AnchorSprite}[2]{\notation{S_{#1}^{#2}}}

\newcommand{\Score}[2]{\notation{s_{#1}^{#2}}}

\newcommand{\SpriteDecoder}{\notation{\mathcal{G}}}
\newcommand{\FrameEncoder}{\notation{\mathcal{E}}}

\newcommand{\Loss}{\notation{\mathcal{L}}}

%% file: include/abstract.tex
Artists and video game designers often construct 2D animations using
libraries of sprites---textured patches of objects and
characters.
We propose a deep learning approach that decomposes sprite-based video
animations into a disentangled representation of recurring graphic
elements in a self-supervised manner.
By jointly learning a dictionary of possibly transparent patches and training a network that places
them onto a canvas, we deconstruct sprite-based content into a sparse,
consistent, and explicit representation that can be easily used in
downstream tasks, like editing or analysis.
Our framework offers a promising approach for discovering recurring visual
patterns in image collections without supervision. 

%% file: include/intro.tex

Since the early days of machine learning, 
the accepted unit of image synthesis has been the {\em pixel}.
But while the pixel grid is a natural representation for display hardware and
convolutional generators, 
it does not easily permit high-level reasoning and editing.  

In this paper, we take inspiration from animation to consider an atomic unit
that is richer and easier to edit than the pixel:  the \emph{sprite}.
In sprite-based animation, a popular early technique for drawing cartoons and
rendering video games, an artist draws a collection of 
patches---a \emph{sprite sheet}---consisting of texture
swatches, characters in various poses, static objects, and so on.
Then, each frame is assembled by compositing a subset of the patches onto a
canvas.
By reusing the sprite sheet, authoring new content requires minimal effort 
and can even be automated procedurally.

Our goal is to invert this process, simultaneously tackling 
unsupervised instance segmentation and dictionary learning.
Given an image dataset, e.g., frames from a sprite-based video game, 
we train a model that jointly learns a 2D sprite dictionary, capturing
recurring visual elements in an image collection, and explains each input
frame as a combination of these potentially transparent sprites.
Whereas standard CNN-based generators hide their feature representation in their
intermediate layers, our model wears its representation ``on its sleeve'':
by explicitly compositing sprites from its learnt dictionary onto a background canvas,
rather than synthesizing pixels from hidden neural features, it provides a
readily-interpretable visual representation.

Our contributions include the following:
\begin{itemize}
    \item We describe a grid-based anchor system along with a learned dictionary of textured patches (with transparency) to extract a sprite-based image representation.
    \item We propose a method to learn the patch dictionary and the grid-based
    representation jointly, in a differentiable, end-to-end fashion.
    \item We compare to past work on learned disentangled graphics representations for video games.
    \item We show how our method offers promising avenues for further work towards identifying visual patterns in more complex data such as natural images and video.
\end{itemize}

%% file: include/related.tex
\section{Related Work}\label{sec:related}

Decomposing visual content into semantically meaningful parts for analysis,
synthesis, and editing is a long-standing problem. We review the most closely
related work.

\textbf{Layered decompositions.}
\citet{wang94layer} decompose videos into  layers undergoing temporally-varying  warps for compression.
Similarly, Flexible Sprites~\citep{jojic2001learning} and
\citet{kannan2005generative} represent videos with full-canvas semi-transparent
layers to facilitate editing.
Like Flexible Sprites, we adopt translation-only
motion but restrict transformations to small neighborhoods around anchors,
making inference tractable with many ($\ge\!100$) sprites. 
Other methods decompose videos with moving subjects, such as humans, into independent layers, enabling matting \cite{lu2021omnimatte}
and retiming of individual actions \cite{lu2020layered}; unlike sprite-based techniques,
motion and appearance are not disentangled.
\citet{sbaio2020pix2vec} use a layered representation as inductive bias in a
GAN with solid colored layers.
Automatic decompositions into ``soft layers'' according to texture,
color, or semantic features have been used in image
editing~\citep{aksoy2017unmixing,aksoy2018semantic}.
\citet{gandelsman2019double} use deep image priors~\citep{ulyanov2018deep} to separate images
into layer pairs. \citet{huang2015efficient} introduce a recurrent architecture to output multiple layers sequentially.
\citet{reddy2020diffcomp} discover patterns in images via differentiable compositing.

\textbf{Interpretable generators for neural synthesis.}
Neural networks improve the fidelity and realism of
generative models~\citep{goodfellow2014gan,karras2020stylegan2} but limit control and
interpretability~\citep{chen2016infogan,harkonen2020ganspace,bau2018gan,bau2019seeing}.
Several works explore interpretability using differentiable domain-specific functions.
\citet{hu2018exposure, li2018differentiable} constrain the generator to sets of 
parametric image operators.
\citet{mildenhall2020nerf} use a  ray-marching prior and rendering model
to encode a radiance field for novel view synthesis.
Neural textures~\citep{thies2019deferred} replace RGB textures on 3D meshes with
high-dimensional features. Rendering under new views  enables view-consistent editing.
\citet{lin2018st} use spatial transformers in their generator to obtain 
geometric transformations.
We synthesize frames by  compositing 2D sprites
undergoing rigid motions, enabling direct interpretation and control over
appearance and motion.

\textbf{Object-centric representations.}
Our learned sprites reveal, segment, and track object instances.
Similarly, Slot Attention~\citep{locatello2020object} extracts object-centric
compositional video representations.
However, our sprites are interpretable---motion and appearance are direct
outputs---and our model scales to more objects per scene.
SCALOR~\citep{jiang2019scalor} handles up to 100 instances but does not
produce a common dictionary or handle diverse sprites.
While SPACE~\citep{Lin2020SPACE} decomposes images into object layers,
it tends to embed sprites in the background, providing no
control.
Our method achieves a higher IoU of recurring
sprite patterns (see \S\ref{sec:comparisons}).
Stampnet~\citep{visser2019stampnet} discovers and localizes objects but
focuses on simpler, synthetic datasets.
MONet~\citep{burgess2019monet} decomposes images into 
multiple object regions using attention.
Earlier attention mechanisms leverage pattern
recurrence~\citep{kosiorek2018sequential,crawford2020exploiting} and motion cues~\citep{eslami2016attend}
to identify individual objects.
Recent works use parametric primitives as image building
blocks~\citep{smirnov2020deep,li2020differentiable}.

Applying our sprite decompositions to video games, we can learn about dynamics
and gameplay, benefiting downstream
agents~\citep{justesen2019deep,he2019tracking} and aiding
content-authoring for research and game development, as in Procedural Content
Generation \citep{summerville2018procedural}.
GameGAN~\citep{Kim2020_GameGan} synthesizes new frames from controller input.
They split rendering into static and dynamic components but 
render full frames, without factorization into parts.
Their generator is difficult to interpret: appearance and dynamics are
entangled within its parameters.

\textbf{Compression.}
Appearance consistency and motion compensation are central to video
compression~\citep{bachu2015videosurvey, lu2018dvc,lombard2019videocompression}.
We model videos as compositions of moving sprites, factoring redundancy in the
input.
This draws inspiration from works like DjVu~\citep{haffner1999djvu} and
Digipaper~\citep{huttenlocher1999digipaper}, which compress scanned documents by
separating them into a background layer and foreground text. 
Image epitomes~\citep{jojic2003epitomes} summarize and compress image shape and
appearance into a miniature texture.
Our sprite dictionary fills a similar role,
providing superior editing control.

%% file: include/method.tex
\input{figs/overview}

\section{Method}\label{sec:method}

We start with an input sequence of $\NumFrames$ RGB frames
$\{\Frame{1}, \ldots, \Frame{\NumFrames}\}$ with resolution $\Width\!\times\!
\Height$.
Our goal is to decompose each frame $\Frame{i} \in \R^{3\times\Width\times
\Height}$ into a set of possibly overlapping sprites, organized into
\NumLayers depth layers, selected from a finite-size dictionary.
The dictionary is a collection of trainable latent codes $\{\DictLatent{1},
\ldots, \DictLatent{\NumPatches}\}$ that are decoded into RGBA sprites using a
neural network generator (\S\ref{sec:dict}).

Our training pipeline is illustrated in Figure~\ref{fig:overview}.
We first process each input frame with a convolutional encoder to produce
\NumLayers grids of feature vectors, one grid per depth layer
(\S\ref{sec:anchors}).
The grids are lower resolution than the input frame, with a
downsampling factor proportional to the sprite size. 
We call the center of each grid cell an \emph{anchor}.
We compare each anchor's feature vector against the dictionary's latent codes,
using a softmax scoring function, to select the best matching sprite per anchor (\S\ref{sec:scoring}).
Using our sprite generator, we decode each anchor's matching sprite. This gives us a grid of
sprites for each of the \NumLayers layers.
To factorize image patterns that may not align with our anchor grid, 
we allow sprites to move in a small neighborhood around anchors
(\S\ref{sec:motion}).
We composite the layers from back to front onto the output canvas to obtain our
final reconstruction (\S\ref{sec:reconstruction}).
Optionally, the background is modeled as a special learnable sprite that covers the entire canvas. 

We train the dictionary latent codes, frame encoder, and sprite generator jointly
on all frames, comparing our reconstruction to the input
(\S\ref{sec:training}).
This self-supervised procedure yields a representation that is sparse,
compact, interpretable, and well-suited for downstream
editing and learning applications.

\subsection{Dictionary and sprite generator}\label{sec:dict}
The central component of our representation is a global \emph{dictionary} of
\NumPatches textured patches or sprites $\Dictionary = \{\Patch{1}, \ldots,
\Patch{\NumPatches}\}$, where each $\Patch{i} \in \R^{4\!\times\!\PatchSize\!\times\!\PatchSize}$ is an
RGBA patch.
Our sprites have an alpha channel, which allows them to be
partially transparent, with possibly irregular (i.e., non-square) boundaries.
This is useful for representing animations with multiple depth layers and also allows to learn
sprites smaller than their maximal resolution, if necessary, by setting alpha to zero 
around the boundary.
The dictionary is shared among all frames; we reconstruct
frames using only sprites from the dictionary.

Instead of optimizing for RGBA pixel values directly, we
represent the dictionary as a set of trainable latent codes $\{\DictLatent{1},
\ldots, \DictLatent{\NumPatches}\}$, with
$\DictLatent{i}\in\mathbb{R}^{\LatentDim}$. We decode these codes into RGBA sprites using a fully-connected
sprite generator $\Patch{i} = \SpriteDecoder\left(\DictLatent{i}\right)$.
This latent representation allows us to define a similarity metric over the
latent space, which we use to pair anchors with dictionary sprites to
best reconstruct the input frame (\S\ref{sec:scoring}).
At test time, we can forego the sprite generator and edit the RGBA sprites
directly.
Unless otherwise specified, we set latent dimension to $\LatentDim=128$ and patch size to $\PatchSize=32$.

We randomly initialize the latent codes from the standard normal distribution.
Our sprite generator first applies zero-mean unit-variance normalization---Layer Normalization \citep{ba2016layer}, 
without an affine transformation---to each
latent code \DictLatent{i} individually, followed by one fully-connected hidden
layer with $8\LatentDim$ features, Group Normalization~\citep{wu2018group}, and
ReLU activation. 
We obtain the final sprite using a fully-connected layer with
sigmoid activation to keep RGBA values in $[0,1]$.
Latent code normalization is crucial to stabilize training and keep
the latent space in a compact subspace as the optimization progresses.
See \S\ref{sec:ablation} for an ablation study of this and other components.

\subsection{Layered frame decomposition using sprite anchors}\label{sec:anchors}

We seek a decomposition 
that best explains each input frame using dictionary sprites.
We exploit translation invariance and locality in our representation;
our sprites are ``attached'' to a regular grid of reference points, or 
\emph{anchors}, inspired by \citep{redmon2016you,girshick2015fast}.
Each anchor has at most one sprite; we call it \emph{inactive} if it has none.

\input{figs/representation}
We give the sprites freedom of motion around their anchors to factorize
structures that may not be aligned with the anchor grid.
This local---or, Eulerian---viewpoint makes inference tractable
and avoids the pitfalls of tracking the 
global motion of
\emph{all} the sprites across the canvas (a Lagrangian viewpoint).
To enable multiple layers with sprite occlusions, we output
$\NumLayers>1$ anchor grids for each frame ($\NumLayers=2$ in our
experiments).
Figure~\ref{fig:representation} illustrates our layered anchor grids and local
sprite transformations.

We use a convolutional encoder \FrameEncoder to map the $\Width\!\times\!\Height$
RGB frame \Frame{i} to \NumLayers grids of anchors, with resolution
$\tfrac{2\Width}{\PatchSize}\!\times\!\tfrac{2\Height}{\PatchSize}$.
Each anchor $j$ in layer $l$ is represented by a feature vector
$\AnchorFeature{j}{l}\in\mathbb{R}^{\LatentDim}$ characterizing 
local image appearance around the anchor and an active/inactive switch
probability $\AnchorProb{j}{l}\in[0,1]$.
%
%
Our frame encoder contains $\log_2(\PatchSize)-1$ downsampling blocks,
which use partial convolutions~\citep{liu2018partialpadding} with kernel size
3 and stride 2 (for downsampling), Group Normalization, and Leaky ReLU.
%
\input{figs/encoder} 
It produces a tensor of intermediate features for each layer, which are
normalized with LayerNorm.
From these, we obtain the anchor switches with an MLP with one hidden layer of
size $\LatentDim$ followed by Group Normalization and Leaky ReLU. We get
anchor features using a linear projection followed by LayerNorm.
The encoder architecture is illustrated in Figure~\ref{fig:encoderarchitecture}.

\subsection{Per-anchor sprite selection}\label{sec:scoring}
Once we have the layered anchor grids for the input frame, we need to assign
sprites to the active anchors.
We do this by scoring every dictionary element $i$ against each anchor $j$ at
layer $l$, using a softmax over dot products between dictionary codes and anchor
features:
%
%
%
\begin{equation}
    \Score{ij}{l} = \tfrac
    {
        \exp\left(
            \nicefrac{\AnchorFeature{j}{l}\cdot\DictLatent{i}}{\sqrt{\LatentDim}}
        \right)
    }
    {
        \sum_{k=1}^{\NumPatches}
        \exp\left(
            \nicefrac{\AnchorFeature{j}{l}\cdot\DictLatent{k}}{\sqrt{\LatentDim}}
        \right)
    }
    .
    \label{eq:score}
\end{equation}
Recall that both the anchor features and dictionary latent codes are
individually normalized using a Layer Normalization operator.
Restricting both latent spaces to a compact subspace helps stabilize
the optimization and avoid getting stuck in local optima.
During training, each anchor's sprite is a weighted combination of the
dictionary elements, masked by the anchor's active probability:
\begin{equation}
    \AnchorSprite{j}{l} = \AnchorProb{j}{l}\sum_{i=1}^{\NumPatches} \Score{ij}{l}\Patch{i}.
\end{equation}
This soft patch selection allows gradients to propagate to
both dictionary and anchor features during training.
Except for natural image and video datasets,
at test time, we use hard selections, i.e., for each anchor, we pick the sprite
($\AnchorSprite{j}{l}:=\Patch{i}$) with highest score \Score{ij}{l} and binarize the switches
$\AnchorProb{j}{l}\in\{0,1\}$.

\subsection{Local sprite transformations}\label{sec:motion}
In real animations, sprites rarely perfectly align with our regular anchor
grid, so, to avoid learning several copies of the same sprites (e.g., all
sub-grid translations of a given image pattern), we allow sprites to move
around their anchors.
In our implementation, we only allow 2D translations of up to
$\nicefrac{1}{2}$ the sprite size on each side of the anchor, i.e.,
$\Transform{j}{l}= (x_j^l, y_j^l) \in [\nicefrac{-\PatchSize}{2},\nicefrac{\PatchSize}{2}]^2$.

We use a convolutional network to predict the
translation offsets from the anchor's sprite and a crop of
the input frame centered around the anchor, with identical spatial dimensions.
This network follows the architecture of $\FrameEncoder$ followed by an MLP
with a single hidden layer of size $\LatentDim$, Group Normalization, and Leaky ReLU.
Specifically, we concatenate the image crop and the anchor's sprite
\AnchorSprite{j}{l} along the channel dimension and pass this
tensor through this network to obtain the $x_j^l$ and $y_j^l$ offsets.
An output layer projects to two dimensions (horziontal and vertical shift) and applies
$\tanh$ to restrict the range.
We apply the shifts to
the patches using a spatial transformer~\citep{jaderberg2015spatial}.
%

%

\input{figs/mario_comparison}

\subsection{Compositing and reconstruction}\label{sec:reconstruction}
Each anchor in our layered representation is now equipped with a sprite
\AnchorSprite{j}{l} and a transformation \Transform{j}{l}.
%
%
For each layer $l$, we transform the sprites in their anchor's local
coordinate system and render them onto the layer's canvas, initialized as
fully transparent.
Because of the local transformation, neighboring sprites within a layer
may overlap.
When this happens, we randomly choose an ordering, as in
Figure~\ref{fig:representation}.
This random permutation encourages our model to either avoid overlapping sprites
within the same layer or 
make the sprite colors agree in the overlap region, since these are the only two
options that yield the same rendering regardless of the random $z$-ordering.
Note that sprites on
\emph{distinct layers} are not shuffled. The shuffling prevents
the network from abusing the
compositing to cover patches with others from the same layer.

We optionally learn a background texture to capture
elements that cannot be explained using sprites.
This can be thought of as a special patch of resolution greater than
that of a single frame. For each frame, we learn a
(discrete) position offset in the background from which to crop.
We represent these offsets as discrete pixel shifts using a softmax classification
(independently for each spatial dimension).
We found this encoding better behaved than using a continuous offset
with a spatial transformer---the discrete encoding allows the gradient signal to
propagate to all shifts rather than the weak local gradient from bilinear
interpolation (see \S\ref{sec:ablation} for an ablation). We combine the background and sprite layers via standard alpha compositing~\citep{porter1984compositing}. Figure~\ref{fig:results_games} shows a learned background.

In some experiments, we use a simpler background model: a fixed solid color,
determined by analyzing the data before training.
In this variant, we sample 100 random frames, cluster the pixel values into 5 clusters
using $k$-means, and choose the largest cluster center as the background color.

\subsection{Training procedure}\label{sec:training}
Our pipeline is fully differentiable.
We train the latent codes dictionary, sprite generator, frame encoder,
transformation predictor, and background layer jointly, minimizing 
$L_2$ distance between our reconstructions and ground truth frames. We also 
employ two regularizers:
a Beta distribution prior on switches and dictionary element scores favors values close to 0 or 1,
and an $L_1$ loss on switches favors a sparser solution without superfluous patches.
Our final loss function for a single input is:

\begin{equation}
    \Loss(\cdot
    ) \!=\! \tfrac1{\Width\Height} \|\Output{}\!-\!\Frame{}\|^2_2 
    \!+\! \tfrac{k^2}{4\NumLayers\Width\Height }\!
    \sum_{l=1}^\NumLayers\!\!\!\sum_{j=1}^{\tfrac{2\Width}{\PatchSize}\!\times\!\tfrac{2\Height}{\PatchSize}}\!\!
    \left[\lambda_\textrm{Beta}\!\left(\!\!\tfrac1\NumPatches\!\sum_{i=1}^\NumPatches \textrm{Beta}(2, 2)(\Score{ij}{l})\!+\!\textrm{Beta}(2, 2)(\AnchorProb{j}{l})\!\!\right)
    \!+\!
    \lambda_\textrm{sparse}|\AnchorProb{j}{l}|\right]\!,\!
\end{equation}
where $\Output{}$ is the result of compositing the background and sprite layers; we optimize $\{s^l_{ij}\}$, $\{p^l_j\}$, and $\Output{}$. We set $\lambda_\textrm{sparse} = 0.005$ and train for 200,000 steps ($\sim\!20$ hours) with $\lambda_\textrm{Beta} = 0.002$ and finetune for 10,000 steps with $\lambda_\textrm{Beta} = 0.1$. For natural images and video, we set $\lambda_\textrm{Beta}=0$.
We use the AdamW \citep{loshchilov2018decoupled} optimizer on a GeForce GTX 1080 GPU, with batch size 4 and learning rate 0.0001, except for the background module (learning rate 0.001 when used).

%% file: figs/overview.tex
\begin{figure*}[ht]
    \centering
    \includegraphics[width=\linewidth]{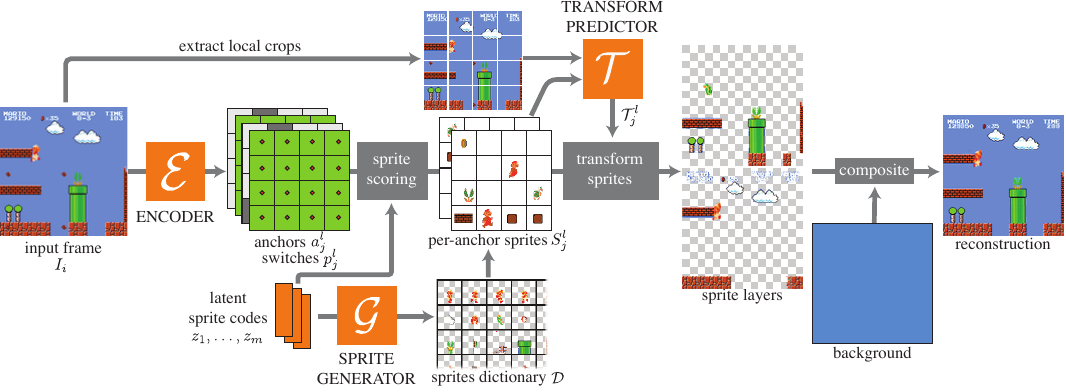}
    \vspace{-0.15in}
    \caption{\label{fig:overview}
        Overview. 
        We jointly learn a sprite dictionary,
        represented as a set of 
        latent codes decoded by a generator, as
        well as an encoder network that embeds a 
        frame into a grid of
        latent codes, or \emph{anchors}.
        Comparing anchor embeddings to dictionary codes
        lets us assign a sprite to each grid cell. 
        Our encoder also outputs a binary switch per anchor to turn sprites on and off.
        After compositing, we obtain a reconstruction of the input. Our self-supervised training optimizes a
        reconstruction loss.
        %
    }
    \vspace{-.2in}
\end{figure*}

%% file: figs/representation.tex
\begin{wrapfigure}[23]{r}{.5\linewidth}
    \centering
    \vspace{-0.15in}
    \includegraphics[width=\linewidth]{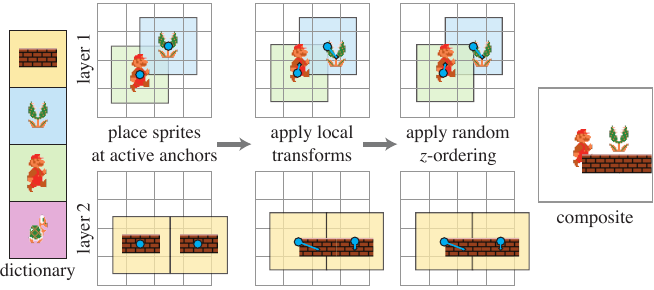}
    \vspace{-0.25in}
    \caption{\label{fig:representation}
        Layered sprite decomposition with local anchors.
        We assign at most one sprite per anchor and
        predict a local transformation of each placed sprite
        around its anchor.
        To allow for occlusions between sprites, we use multiple sprite layers,
        which we compose back to front to obtain the final image.
 }
 \vspace{0.15in}\par
\includegraphics[width=\linewidth]{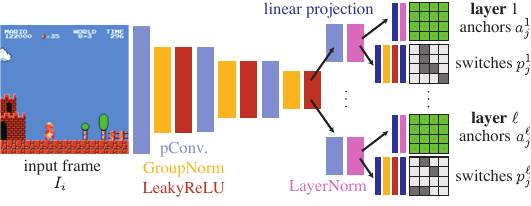}
\vspace{-.3in}
\caption{Encoder architecture.\label{fig:encoderarchitecture}}
\end{wrapfigure}

%% file: figs/encoder.tex

%% file: figs/mario_comparison.tex
\begin{figure*}[!t]
    \centering
    \includegraphics[width=\linewidth]{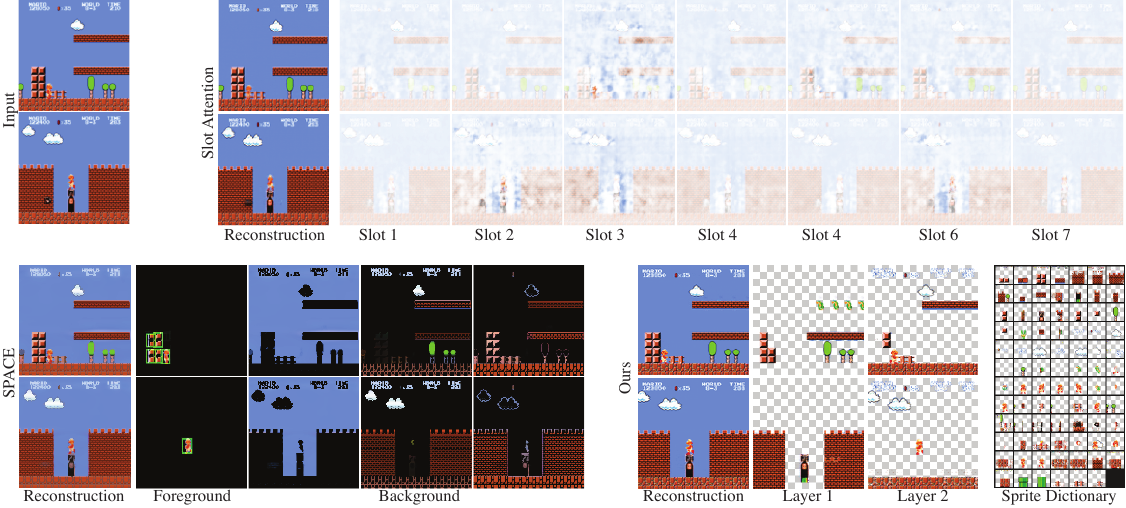}
    \vspace{-0.2in}
    \caption{\label{fig:mario_comparison}
        {Comparison to SPACE \citep{Lin2020SPACE} and Slot Attention \citep{locatello2020object}. While all three methods obtain good reconstructions, SPACE only recognizes a few sprites, and Slot Attention does not yield a meaningful decomposition. We decompose the entire foreground and learn a dictionary.}
    }
    \vspace{-0.1in}
\end{figure*}

%% file: include/results.tex
\section{Experimental Results}\label{sec:results}

We evaluate our self-supervised decomposition on several real (non-synthetic) datasets, compare to related work, and conduct an ablation study. 
In figures, we use a checkerboard to show transparency.
Dictionary order is determined by sorting along a 1-dimensional $t$-SNE embedding
of the sprite latent codes.
We find this ordering tends to group 
semantically similar sprites, making the dictionary easier to interpret and
manipulate. While our models are trained with a dictionary of 150
patches, not all patches end up being used; we only
show the used patches.

\subsection{Comparisons}
\label{sec:comparisons}

\input{figs/synth_comparison}

While to our knowledge no prior works target
differentiable unsupervised sprite-based reconstruction, we compare to two
 state-of-the-art methods that obtain similarly disentangled
representations.

In Figure~\ref{fig:mario_comparison}, we compare to SPACE~\citep{Lin2020SPACE}
and Slot Attention~\citep{locatello2020object}.
The former decomposes a scene
into a foreground layer consisting of several objects as well as a background,
segmented into three layers.
The latter deconstructs a scene into 
discrete ``slots.''
We train both methods to convergence using their default
parameters.
While both reconstruct the input frames faithfully, SPACE
only recognizes a few sprites in its foreground layer, and Slot Attention does
not provide a semantically meaningful decomposition. 
In contrast, not only does our method model the entire scene using learned sprites, but also it
factors out the sprites to form a consistent, sparse dictionary shared for the
entire sequence.

\input{tables/space}

Additionally, we evaluate on a synthetically-generated sprite-based game from
\citep{dubeyICLRW18human}, which is made of sprites on a solid background. We
compare quantitatively to SPACE in 
Figure~\ref{tab:space} and show qualitative results in Figure~\ref{fig:synth_comparison}. Since we have a ground truth segmentation of each scene
into sprites, we compute a matching between learned dictionary patches and sprites by
associating each dictionary patch with the sprite that it most frequently overlaps
across the dataset. We visualize dictionary patches next to their respective sprites.
We also use this labeling to compute segmentation metrics.
In particular, we report mean IoU in the multiclass case (where each sprite is a
distinct class) as well is in the binary case (foreground/background).
Because SPACE does not learn a common dictionary, we are unable to
obtain a labeling for its foreground elements and, consequently, cannot
evaluate
its multiclass metric. For the binary metric, we obtain a significantly
higher value, since SPACE  defers many sprites to the background, whereas
our method learns the sprites as dictionary elements.

\input{figs/baselines}

To show that our model learns more than simple motion features,
we also compare to two conventional (non-learning) baselines. In Figure~\ref{fig:baselines}(a), we compare a segmentation of a frame obtained by clustering optical flow directions using $k$-means (inspired by \citet{liu2005motion}) to one generated using our learned decomposition. The flow-based approach is unable to capture many of the details in the frame. In (b), we show the normalized dictionary obtained using an online dictionary learning method~\citep{mairal2010online}. Because this method does not have the inductive biases of our model, the resulting dictionary is not easily interpretable or editable.

\subsection{Sprite-based game deconstruction}

We train on Fighting Hero (one level, 5,330 frames), Nintendo Super Mario Bros.\ (one level, 2,220 frames), and ATARI Space Invaders (5,000 frames). We use patch size $\PatchSize=32$ for Mario and Fighting Hero and $\PatchSize=16$ for Space Invaders. For Fighting Hero, we learn a background, as described in \S\ref{sec:reconstruction}.

\input{figs/results_games}

\input{figs/mario_edit}

The sprites, background, and example frame reconstructions are shown in Figure~\ref{fig:results_games}. Our model successfully disentangles foreground from background and recovers a reasonable sprite sheet for each game. Having reverse-engineered the games, we can use the decomposition for applications like editing. In Figure~\ref{fig:mario_edit}, we demonstrate a GUI that allows the user to move sprites around the screen.

\subsection{Ablation Study}\label{sec:ablation} 

\input{tables/ablation}

We show an ablation study on the Mario data. We train our full model, one with smaller $16\!\times\!16$ patches, another with larger $64\!\times\!64$ patches, a model with a smaller dictionary (25 elements), a model without LayerNorm, and one where we use a straight-through estimator \citep{jang2016categorical} to learn discrete switches $\AnchorProb{j}{l}$ in lieu of Beta regularization. We train each model with five random seeds and report the reconstruction PSNR means and standard deviations in Figure~\ref{tab:ablation}. This experiment verifies the importance of LayerNorm in our architecture and shows that the straight-through trick is ineffective in our setting. Though the smaller patches model achieves slightly higher mean PSNR than our full model, more of the sprites are split across dictionary patches (Figure~\ref{fig:smaller_patches}), illustrating how the patch size choice sets an inductive bias for our decomposition.

\input{figs/bg_st}

We also justify our choice for learning background shifts via classification (\S\ref{sec:reconstruction}) rather than regression, i.e., using spatial transformers. Figure~\ref{fig:bg_st} shows the background learned using a spatial transformer. In contrast to our full model (Figure~\ref{fig:results_games}), the original background is not discovered, and most of the canvas is unused. We suspect that this is due to lack of gradient signal from background pixels that do not get rendered at each training step.

\subsection{Future Directions and Limitations}\label{sec:limitations}

\input{figs/results_compression}

While our method is designed with sprite-based animation in mind, it can generalize to natural images and videos. An exciting direction for future work is to incorporate more expressive transformations so as to discover recurring content in generic videos. Here, we obtain preliminary results using our approach and achieve interesting decompositions even without modifications to our sprite-based model.

\input{figs/results_tennis}

In Figure~\ref{fig:results_tennis}, we show results on a tennis video (4,000 frames). The model learns parts of the player's body (head, limbs, shirt, etc.)
as sprites and captures most of the tennis court in the learned background. By simply selecting the player sprites in the dictionary,
we  segment the entire video clip.

Our model can also discover recurring patterns in a single natural image. We train on random crops of a $768\!\times\!512$ photograph from the 2013
US Presidential Inauguration,\footnote{AP Photo/Cliff Owen} which contains many repeating elements such as stairs, columns, and people. With a dictionary of 39 $32\!\times\!32$ sprites (39,936 pixels),
we recover much of the detail of the original 393,216 pixels.

\input{figs/results_text}

We demonstrate further limitations of our approach by applying it to automatic font discovery.
We train on random $128\!\times\!128$ crops of six scanned pages of \emph{Moby Dick}, each of approximately $500\!\times\!800$ resolution.
Figure~\ref{fig:results_text} shows an input text excerpt, our reconstruction, and the learned dictionary.

This dataset differs significantly from our other testing datasets. Each input frame consists of many densely packed sprites ($\sim\!100$ glyphs in each $128\!\times\!128$ crop), and
many individual glyphs consist of smaller repeating elements.
We hypothesize that because of these issues, combined with a lack of motion cues between frames, we do not achieve a perfect reconstruction, learning certain sprites with multiple glyphs and others with just partial glyphs.
Incorporating priors tailored to regularly structured and dense data like text is a direction for future research.

%% file: figs/synth_comparison.tex
\begin{figure}[!tb]
    \centering
    \includegraphics[width=\linewidth]{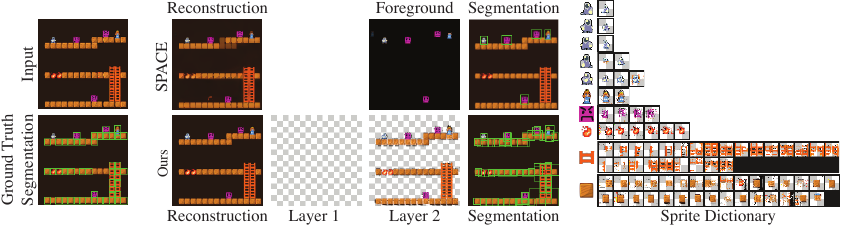}
    \vspace{-0.2in}
    \caption{\label{fig:synth_comparison}
        {Qualitative comparison to SPACE \cite{Lin2020SPACE} on the synthetic game dataset. We show ground truth sprite segmentations as well as those obtained from the learned SPACE foreground and from our learned sprites. While SPACE only learns several of the sprites, we reconstruct the entire foreground using our dictionary.}
    }\vspace{-.15in}
\end{figure}

%% file: tables/space.tex
\begin{figure}[b!]
    \centering
    \scriptsize
    \begin{tabular}{ccccc}
         \toprule
         Method & Reconstruction PSNR & Mean IoU (multiclass) & Mean IoU (binary)\\
         \midrule
         SPACE & 31.9 & - & 0.0361\\
         Ours & \bf 38.54 & \bf 0.6497 & \bf 0.7352\\
         \bottomrule
    \end{tabular}
    \caption{Quantitative comparison to SPACE \citep{Lin2020SPACE}. We report PSNR to evaluate overall reconstruction quality as well as mean IoU for multiclass and binary foreground/background segmentation problems. Our method recognizes significantly more sprites than SPACE, resulting in higher mean IoU.}
    \label{tab:space}
\end{figure}

%% file: figs/baselines.tex
\begin{wrapfigure}[10]{r}{0.3\linewidth}
    \centering
    \vspace{-0.25in}
    \includegraphics[width=\linewidth]{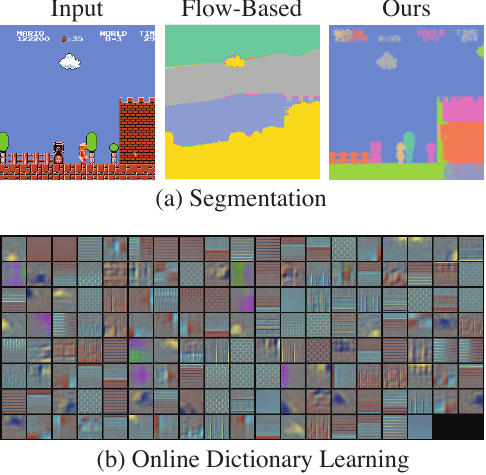}
    \vspace{-0.25in}
    \caption{\label{fig:baselines}
        {Comparison to conventional baselines.}
    }
\end{wrapfigure}

%% file: figs/results_games.tex
\begin{figure}[t]
    \centering
    \includegraphics[width=\linewidth]{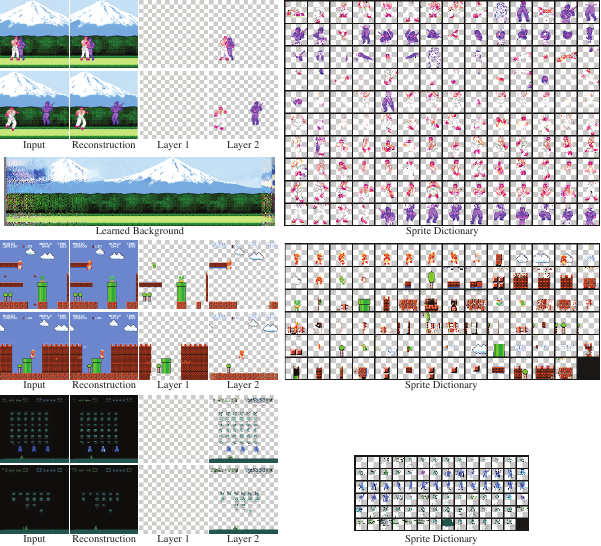}
    \vspace{-0.25in}
    \caption{\label{fig:results_games}
        Sprite-based game decompositions.
        Our self-supervised technique recovers a compact dictionary of
        semantically meaningful sprites representing characters (or their body
        parts) and props.
        The first example shows our learned background texture; the others
        use a solid color as background.
    }\vspace{-.15in}
\end{figure}

%% file: figs/mario_edit.tex
\begin{wrapfigure}[5]{r}{.3\linewidth}
    \centering
    \vspace{-0.1in}
    \includegraphics[width=\linewidth]{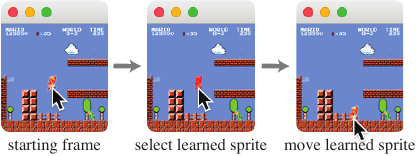}
    \vspace{-0.2in}
    \caption{\label{fig:mario_edit}
        {Editing GUI.}
    }
\end{wrapfigure}

%% file: tables/ablation.tex
\begin{wrapfigure}[14]{r}{.35\linewidth}
    \centering
    \vspace{-0.3in}
    \scriptsize
    \begin{tabular}{cc}
         \toprule
         Model & PSNR \\
         \midrule
         Smaller patches & $28.85 \pm 0.95$ \\
         Full & $28.04 \pm 0.72$ \\
         No LayerNorm & $26.05 \pm 0.45$ \\
         Smaller dictionary & $23.80 \pm 1.38$ \\
         Larger patches & $23.63 \pm 1.05$ \\
         Straight-through switches & $22.15 \pm 0.25$ \\
         \bottomrule
    \end{tabular}
    \vspace{-0.1in}
    \caption{\label{tab:ablation}
    {Ablation study on Mario data across five random seeds.}
    }
    
    \vspace{0.1in}

    \includegraphics[width=\linewidth]{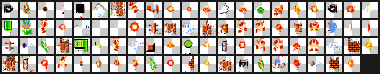}
    \vspace{-0.2in}
    \caption{\label{fig:smaller_patches}
        Mario dictionary with $16\!\times\!16$ patches.
    }
\end{wrapfigure}

%% file: figs/bg_st.tex
\begin{wrapfigure}[6]{r}{0.35\linewidth}
    \centering
    \vspace{-0.15in}
    \includegraphics[width=\linewidth]{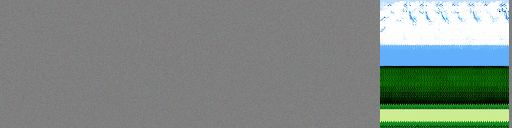}
    \vspace{-0.25in}
    \caption{\label{fig:bg_st}
        {Background learned with spatial transformer.}
    }
\end{wrapfigure}

%% file: figs/results_compression.tex
\begin{wrapfigure}[8]{r}{.5\linewidth}
    \centering
    \vspace{-0.1in}
    \includegraphics[width=\linewidth]{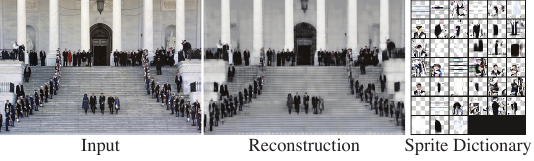}
    \vspace{-0.25in}
    \caption{\label{fig:results_compression}
        We factorize the recurring elements of a single photograph into a compact dictionary.
    }
\end{wrapfigure}

%% file: figs/results_tennis.tex
\begin{figure*}[ht]
    \centering
    \includegraphics[width=\linewidth]{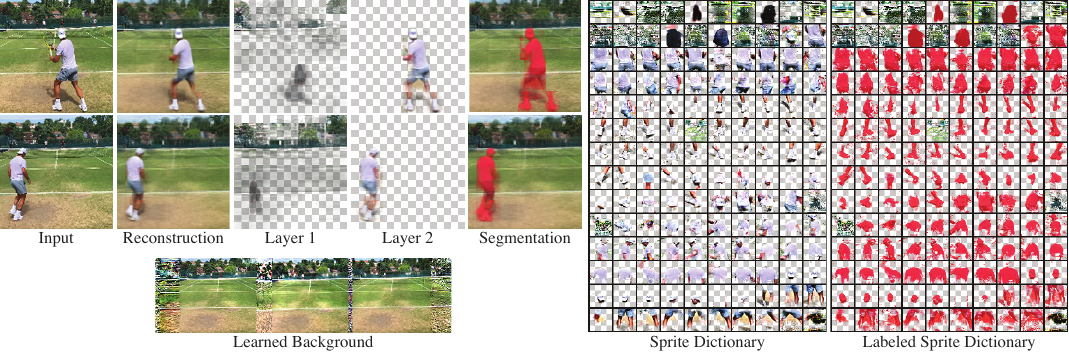}
    \vspace{-0.2in}
    \caption{\label{fig:results_tennis}
        Segmentation of natural videos.
        Despite its simplistic motion and appearance model, our
        approach can be applied to real-world videos.
        By selecting the few sprites corresponding to the tennis player,
        we can quickly obtain a segmentation of the full video sequence.
    }
    \vspace{0.1in}
\end{figure*}

%% file: figs/results_text.tex
\begin{figure}[t!]
    \centering
    \includegraphics[width=\linewidth]{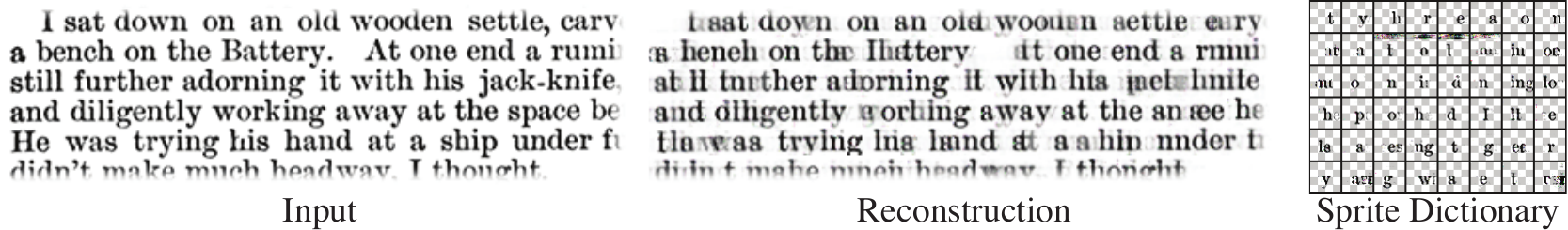}
    \vspace{-0.2in}
    \caption{\label{fig:results_text}
        Reconstruction of a scanned text excerpt, illustrating some limitations of our method.
        Because letters are densely packed within the text and lack motion cues, our model learns sprites comprising more than a single glyph
        and suffers from high reconstruction error.
    }
\end{figure}

%% file: include/conclusion.tex
\section{Conclusion}

We present a self-supervised method to jointly learn a patch dictionary and a
frame encoder from a video, where the encoder explains frames as compositions of
dictionary elements, anchored on a regular grid.
By generating layers of alpha-masked sprites and predicting per-sprite
local transformation, we recover fine-scale motion and achieve
high-quality reconstructions with semantically meaningful,
well-separated sprites.
Applied to content with significant recurrence, our approach recovers
structurally significant patterns.
 
Understanding recurring patterns and their relationships is central to machine
learning.
Learning to act intelligently in video games or in the physical world
requires breaking experiences down into elements between which
knowledge can be transferred effectively. Our sprite-based
decomposition provides an intuitive basis for this purpose.
In this work, we focus on a simplified video domain. 
In the future, we would like to
expand the range of deformations applied to the learned dictionary
elements, such as appearance or shape changes.
Our work opens significant avenues for future research to explore
recurrences and object relationships in more complex domains.

%% file: include/acknowledgements.tex
\section*{Acknowledgements}
The MIT Geometric Data Processing group acknowledges the generous support of Army Research Office grants W911NF2010168 and W911NF2110293, of Air Force Office of Scientific Research award FA9550-19-1-031, of National Science Foundation grants IIS-1838071 and CHS-1955697, from the CSAIL Systems that Learn program, from the MIT--IBM Watson AI Laboratory, from the Toyota--CSAIL Joint Research Center, from a gift from Adobe Systems, from an MIT.nano Immersion Lab/NCSOFT Gaming Program seed grant, and from the Skoltech--MIT Next Generation Program. This work was also supported by the National Science Foundation Graduate Research Fellowship under Grant No.\ 1122374.